\documentclass{article} 
\usepackage[final]{colm2026_conference}

\usepackage{microtype}
\usepackage{hyperref}
\usepackage{url}
\usepackage{booktabs}
\usepackage{graphicx}
\usepackage{subcaption}

\usepackage{amsmath}
\usepackage{amssymb}
\usepackage{mathtools}
\usepackage{amsthm}
\usepackage{wrapfig}
\usepackage[capitalize,noabbrev]{cleveref}

\usepackage{xcolor}
\usepackage[most]{tcolorbox}
\definecolor{googleblue}   {HTML}{4285F4}
\definecolor{googlered}    {HTML}{EA4335}
\definecolor{googleyellow} {HTML}{FBBC05}
\definecolor{googlegreen}  {HTML}{34A853}
\definecolor{googlegray}      {HTML}{5F6368}

\usepackage{enumitem}
\usepackage[table]{xcolor}
\usepackage{balance}  
\usepackage{multirow}
\usepackage{microtype} 

\usepackage{subcaption}
\usepackage{pifont}
\newcommand{\cmark}{\ding{51}}
\newcommand{\xmark}{\ding{55}}
\definecolor{cvprblue}{rgb}{0.21,0.49,0.74}

\usepackage{lineno}

\definecolor{darkblue}{rgb}{0, 0, 0.5}
\hypersetup{colorlinks=true, citecolor=darkblue, linkcolor=darkblue, urlcolor=darkblue}

\title{SVR-R1: Bootstrapping Multi-modal Reasoning with Self-verification in Reinforcement Learning}

\author{%
Mingyuan Wu$^{1}$,
Jingcheng Yang$^{1}$,
Shengyi Qian$^{2}$,
Xudong Wang$^{2}$,
Jize Jiang$^{1}$, \\
\bf Qifan Wang$^{2}$,
\bf Aashu Singh$^{2}$,
\bf Khoi Pham$^{2}$,
\bf Fei Liu$^{2}$,
\bf Zhaolun Su$^{2}$, \\
\bf Zhuokai Zhao$^{2}$,
\bf Klara Nahrstedt$^{1}$,
\bf \bf Jianyu Wang$^{2}$,
Hanchao Yu$^{2}$ \\
$^{1}$University of Illinois Urbana-Champaign,
$^{2}$Meta \\
\texttt{\{mw34\}@illinois.edu} \\
}

\begin{document}

\ifcolmsubmission
\linenumbers
\fi

\maketitle

\begin{abstract}
We introduce \textbf{Self-Verified Reasoner (SVR-R1)}, a multi-turn RL framework that turns a model’s own verification into a learning signal for multimodal reasoning. For each query, the model proposes an answer using the same weights, and issues a binary self-verdict (Yes/No). A “No” triggers a second-chance rethink; a “Yes,” or a turn cap, finalizes the output for computing the outcome-based reward. \textbf{SVR-R1} is implemented with GRPO and an asynchronous multi-turn rollout framework and needs no external supervision or auxiliary critics. We evaluate \textbf{SVR-R1} on vision-language reasoning benchmarks and show that it improves accuracy by a large margin over strong standard GRPO baselines. Training dynamics show decreasing reliance on verification-fewer verification turns, yet higher test accuracy-indicating that the gap between verification and generation narrows as the policy internalizes self-correction and chooses the most confident answer via our framework. \textbf{SVR-R1} bridges the less explored intersection of inference-time self-refinement and RL training for VLMs, offering a simple yet effective recipe for bootstrapping multimodal reasoning. We will open-source \textbf{SVR-R1} to facilitate future research in VLMs.
\end{abstract}
\section{Introduction}

\label{sec:intro}
\begin{wrapfigure}{r}{0.53\columnwidth}
  \centering
  \vspace{-2.8\baselineskip}
  \includegraphics[width=\linewidth]{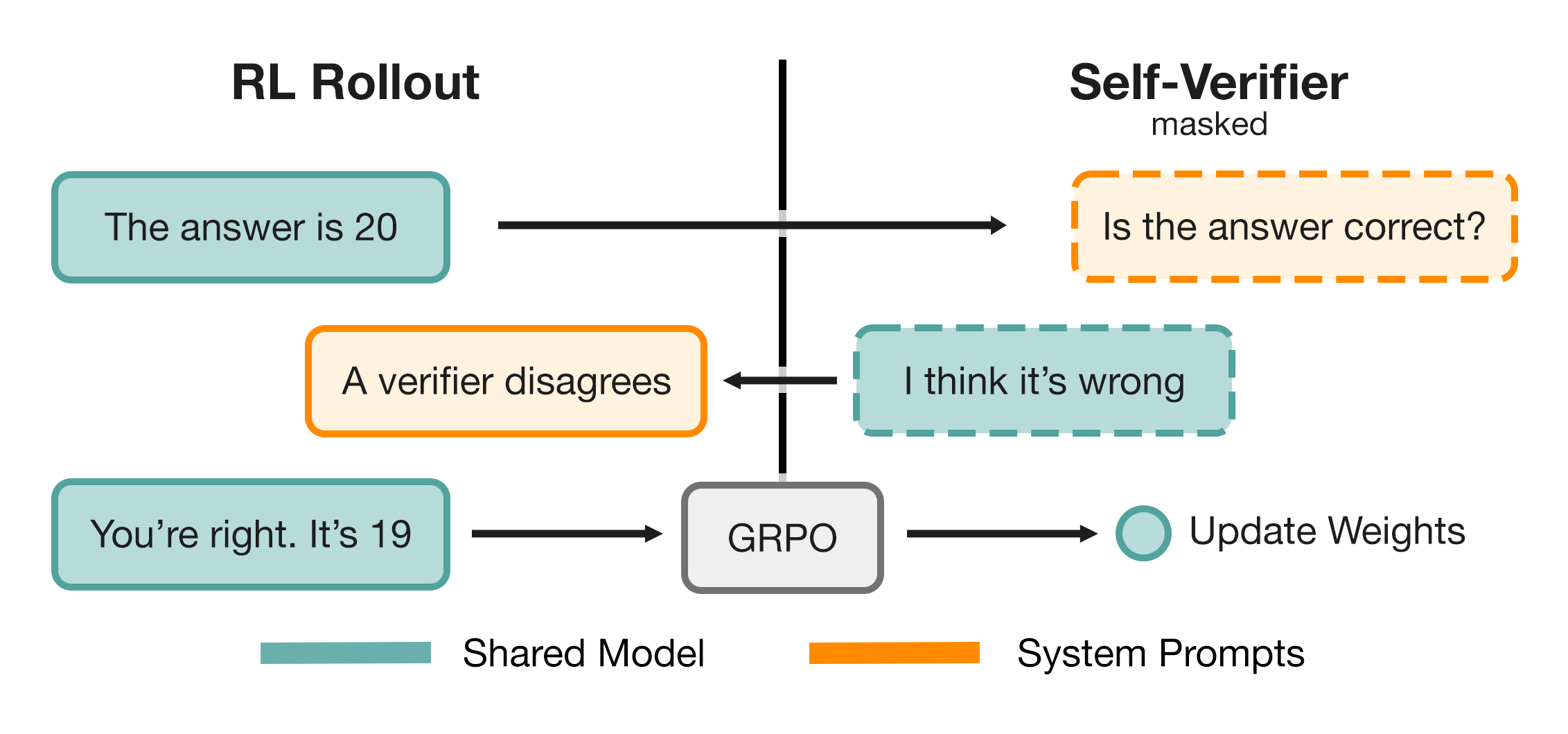}
  \vspace{-1.5\baselineskip}
  \caption{Merging VLM Self-Verification Loop into RL Rollout}
  \label{fig:thumb}
  \vspace{-1.0\baselineskip}
\end{wrapfigure}
When given a second chance to think, humans often reason their way toward better solutions on complex problems. Recently, Large Language Models (LLMs) have demonstrated a similar pattern of iterative rethinking in the reasoning process~\citep{openai2024openaio1card}, particularly when fine-tuned with reinforcement learning (RL) on task-specific rewards~\citep{R1, zeng2025simplerlzooinvestigatingtamingzero, zhou2025mixture, wang2025sotalessmctsguidedsample}. After such training, these models can intrinsically exhibit reasoning behaviors that allow them to iteratively improve their reasoning paths through behaviors like self-correction and backtracking~\citep{gandhi2025cognitivebehaviorsenableselfimproving}. This approach was initially applied to verifiable tasks in the language domain, such as mathematics and coding, and has since been extended to non-verifiable domains and to vision-language tasks~\citep{zhou2025r1zerosahamomentvisual, chen2025r1v, zhang2025r1vllearningreasonmultimodal, huang2025visionr1incentivizingreasoningcapability, deng2025openvlthinker, wang2025vlrethinkerincentivizingselfreflectionvisionlanguage}.

Even before the widespread adoption of RL fine-tuning for enabling self-rethinking in LLMs~\citep{kumar2025training}, it was observed that simply prompting an LLM to second-guess to refine or correct its previous answer could lead to improved reasoning performance in inference~\citep{madaan2023selfrefineiterativerefinementselffeedback, reflecxion}. This improvement may be attributed to the gap between verification and generation~\citep{song2025mindgapexaminingselfimprovement}: \textit{``if verification is easier than generation, one can hypothesize that the model may act as a better-than-random verifier of its own outputs, enabling
self-improvement''}. Beyond pure language, pioneering work~\citep{vlm-self-verify} has shown that strong commercial Vision Language Models (VLMs) such as GPT-4o can, to some extent, self-verify their outputs in certain tasks (e.g. segmentation), though effective self-rethinking generally remains challenging and underexplored in multi-modal reasoning scenarios~\cite{wu2025ahamomentrevisitedvlms, huang2025autonomous}.

In this work, we explore a previously unexplored middle ground between explicit prompting and RL fine-tuning for self-rethinking by leveraging VLMs' pre-existing self-verification capabilities in Figure~\ref{fig:thumb}. Specifically, we aim to bootstrap a VLM's reasoning abilities, prompting it to rethink if the model itself deems its (tentative) answer wrong, and finalizes its output if it deems its answer correct. To accomplish this, we integrate model self-verification rounds, \textit{which share the same copy of model weights}, into the RL training process prior to reward assignment. Specifically, we interleave multi-turn self-generation and verification during RL rollouts: the self-verifier is presented with the question and the initial self-generated solution, and is prompted to provide a binary verification of correctness (\textit{Yes or No}). If the answer is \textit{No}, the model is given a second chance to think and regenerate its answer accordingly. If the answer is \textit{Yes}, or if a manually set maximum number of verification rounds is reached, the model’s final response is passed to reward calculation. The overall training is conducted using the widely adopted Group Relative Policy Optimization algorithm (GRPO)~\citep{grpo} and outcome-based reward matching, and supported by asynchronous multi-modal multi-turn RL framework~\citep{sheng2024hybridflow} to coordinate the self-verifier and generator in the rollout. We refer to our approach as the \textbf{Self-Verified Reasoner (SVR-R1)}.

\begin{wrapfigure}{r}{0.5\columnwidth}
  \centering
  \vspace{-1.8\baselineskip}
  \includegraphics[width=\linewidth]{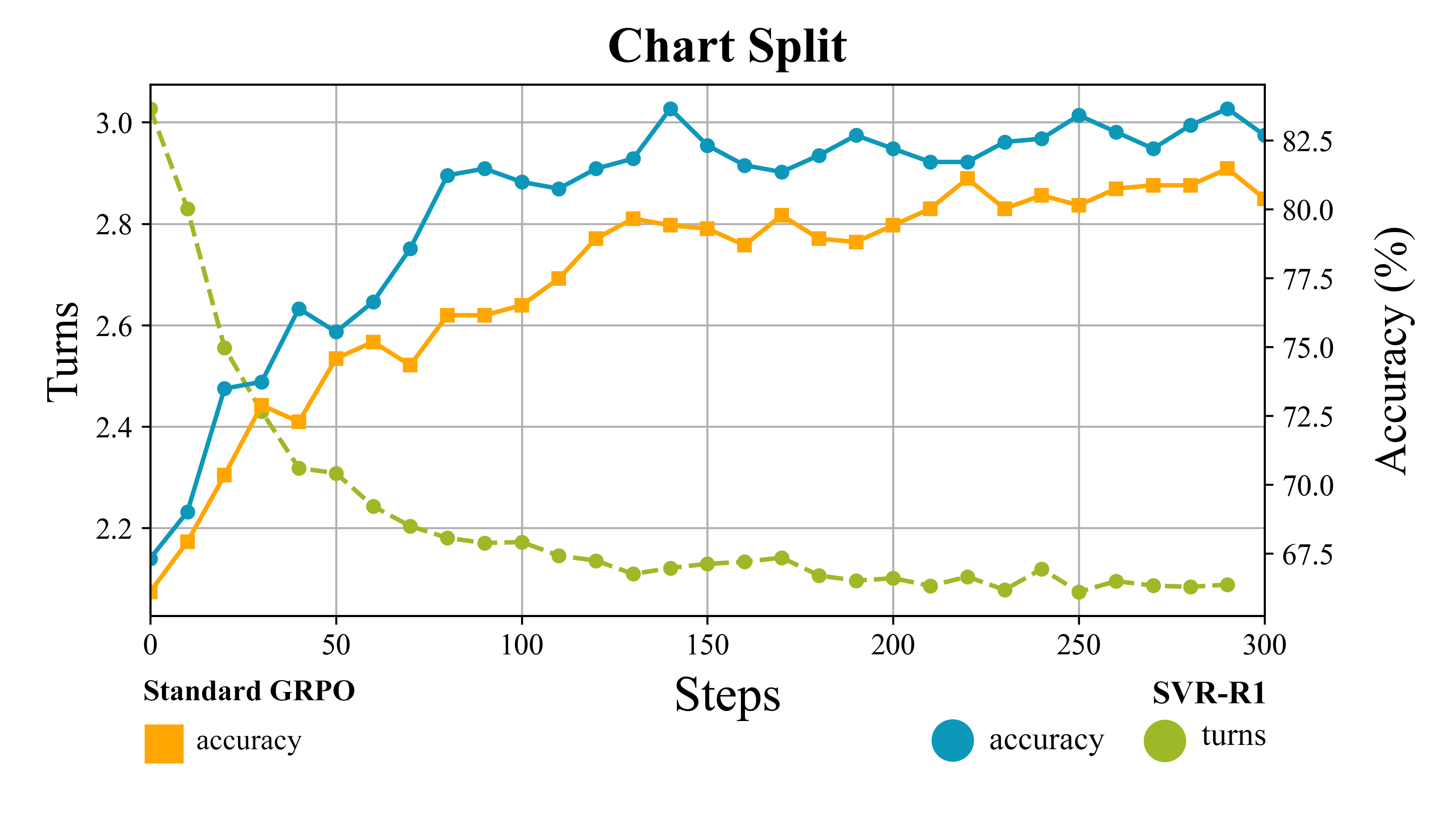}
  \vspace{-2em}
  \caption{\textbf{Validation Reasoning Accuracy (\%)} vs. \textbf{Training Steps}. SVR-R1 compared with standard GRPO~\citep{grpo} on the Qwen2.5-VL~\citep{qwenvl25} 3B model, with \textbf{Mean number of Turns} decreasing.}
  \label{fig:thumb-acc-1col}
  \vspace{-1\baselineskip}
\end{wrapfigure}

We demonstrate the effectiveness of \textbf{SVR-R1} across multiple challenging multi-modal table and chart reasoning benchmarks~\citep{fu2025refocus}, as well as general reasoning tasks~\citep{wang2025sotalessmctsguidedsample},  showing significant improvements in vision-language reasoning performance. Interestingly, we observe that, over the course of training, the policy model gradually performs fewer verification rounds, eventually leading to the self-verifier almost always immediately affirming the answer in a single round, while still achieving improved accuracy on the test set, as it is demonstrated in Figure~\ref{fig:thumb-acc-1col}. This suggests that \textbf{SVR-R1} helps the model learn to close the gap between generation and verification by exploiting self-verification during RL training.

\textbf{SVR-R1} also contributes to the broader direction of self-improvement without external supervision. While most early self-improvement approaches~\citep{huang-etal-2023-large, star, zhao2025boosting} have focused on using self-generated, high-confidence reasoning traces for supervised fine-tuning, they have not fully integrated these traces into the iterative RL pipeline. In this regard, \textbf{SVR-R1} advances the understanding of how models can continuously bootstrap themselves through interleaved generation and verification in RL training, without relying on extra external data or model supervision.
\section{Related Work}

\subsection{RL for LLM and VLM Reasoning}
Reinforcement learning (RL) became widely adopted for LLM development through RL from Human Feedback (RLHF)~\citep{ouyang2022traininglanguagemodelsfollow}, which uses a reward model trained on human preferences to optimize the LLM policy via Proximal Policy Optimization (PPO)~\citep{schulman2017proximal}. More recent work has introduced computationally efficient variants of PPO~\citep{rafailov2023direct, wang2024preference, grpo, zhou2025disco}, making RL-based training more accessible. 

Beyond alignment on human preference, RL has demonstrated significant improvements in LLM reasoning and self-correction capabilities~\citep{kumar2025training, R1, zeng2025simplerlzooinvestigatingtamingzero}. Recent studies have investigated the intrinsic properties that enable effective self-improvement, revealing emergent behaviors such as "aha moments" that arise through RL-based fine-tuning~\citep{gandhi2025cognitivebehaviorsenableselfimproving, zeng2025simplerlzooinvestigatingtamingzero}. 
Building on these advances, RL fine-tuning has also been successfully adopted for multimodal reasoning tasks~\citep{zhou2025r1zerosahamomentvisual, chen2025r1v, zhang2025r1vllearningreasonmultimodal, huang2025visionr1incentivizingreasoningcapability, liu2025segzeroreasoningchainguidedsegmentation, deng2025openvlthinker, wang2025vlrethinkerincentivizingselfreflectionvisionlanguage, wang2025sotalessmctsguidedsample}. Notably, VL-Rethinker~\citep{wang2025vlrethinkerincentivizingselfreflectionvisionlanguage} elicits self-reflection in vision–language models through prompting. In contrast, our work is the first to integrate self-reflection directly into the RL post-training process.

\subsection{Self-Improvement of LLMs/VLMs}
Early work explored inference- or prompt-level self-improvement, such as asking the model to generate feedback and revise its responses, or manually incorporating re-verification into the prompt~\citep{madaan2023selfrefineiterativerefinementselffeedback, weng2023large}. Other methods \citep{huang-etal-2023-large, star, zhao2025boosting, zhou2025mixture} encourage models to generate reasoning traces, filter high-quality responses, and then fine-tune on these traces to achieve model self-improvement via next-token prediction. Subsequent studies attribute such self-improvements to mechanisms such as sharpening~\citep{sharpen} or to the verification–generation gap in reasoning tasks~\citep{song2025mindgapexaminingselfimprovement}, and try merging with the RL workstreams~\citep{jiang2025bootstrappingtaskspacesselfimprovement, chen2025scaling}. 

Recently, efforts to enable self-improvement have expanded into the vision-language domain~\citep{vlm-self-verify, wu2025ahamomentrevisitedvlms, ding2025sherlockselfcorrectingreasoningvisionlanguage}, though these studies are typically limited to constrained settings. Progress in this area remains challenging, likely due to difficulties in multi-modal pretraining and lack of high quality reasoning data compared to pure language setting.
\section{Method}
This section describes \textbf{SVR-R1}'s bootstraping of VLMs' reasoning abilities: \textbf{SVR-R1} aims to enable VLMs to reach final reasoning answers that the model \textit{itself} deems correct via guiding the model to re-iterate on answers that the model deems wrong.
We describe \textbf{SVR-R1}'s overall pipeline in~\cref{sec:method_preliminary}, self-generator and verifier protocol in~\cref{sec:method_protocol}, and accordingly, RL-based multi-turn self-generation and verification in~\cref{sec:method_rl}, with learning objectives and outcome-based reward design.

\begin{figure*}[t]
    \centering
   \includegraphics[width=\linewidth]{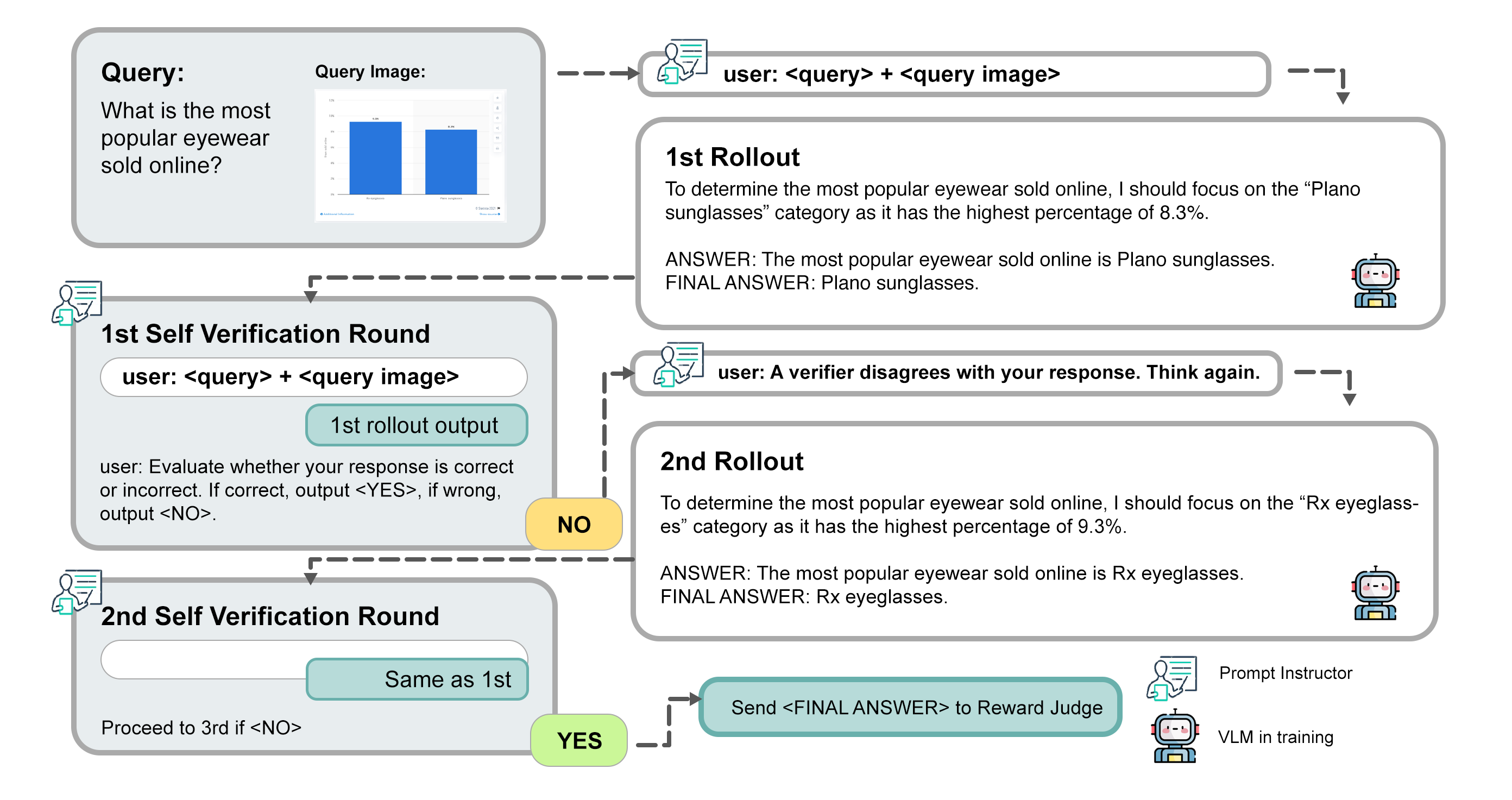}
   \vspace{-2em}
    \caption{Multi-rollout with self-verification. Each rollout, along with additional prompts, \textbf{becomes part of the context} for the next.}
    \vspace{-1.5em}
    \label{fig:method}
\end{figure*}

\subsection{\textbf{SVR-R1} Overview}
\label{sec:method_preliminary}
This section describes how \textbf{SVR-R1}, as a VLM policy, is designed to process and respond to multi-modal prompts.

\noindent \textbf{Preliminaries.} We denote a VLM policy as $\pi_{\theta}$ parametrized with model weights $\theta$ in this paper. Given a text prompt sequence $x$ and an image $I$, the model can generate a text response sequence $y$, sampled from the $\pi_{\theta}(I, x)$. The weight $\theta$ is shared by \textbf{SVR-R1}'s self-generator and verifier during RL rollouts (\cref{sec:method_rl}) despite operating with different input text instruction prompts (\cref{sec:method_protocol}).

\noindent \textbf{SVR-R1 Pipeline.}
We construct the initial prompt headers to be passed to \textbf{SVR-R1} to include carefully designed requirements and few-shot examples that can enhance multimodal reasoning (prompt in Appenedix \cref{fig:tcolorbox-vlm-head}), inspired by the approach of previous work \citep{fu2025refocus}.
Then, \textbf{SVR-R1}'s multi-turn conversation for self-generation and verification proceeds as an interleaving sequence of user turns (for model prompt input) and assistant turns (for model output), following the standard conversational templates commonly used in VLMs such as Qwen series \citep{qwenvl25}:
In the assistant turn, the model generates a reasoning response that includes both its rationale and answer, which is then passed to the following self-verification step. All user and assistant turns are preserved in the conversation history as the model progresses through successive rounds of generation and verification towards the outcome-based judge. An overview of the entire SVR-R1 pipeline is depicted in \cref{fig:method}, where we visualize the different model inputs, outputs, and conversational turns in distinct colors, with all elements embedded within the history.

\subsection{Self-Verification and Generation}
\label{sec:method_protocol}
This section describes \textbf{SVR-R1}'s self-verifier and generators, which share the same set of model weights $\theta$ during RL rollouts but notably operate with different input text prompts. \textbf{SVR-R1} enforces interleaving of self-verification and generation steps, repeating this process until the maximum number of rounds is reached.

\begin{figure*}[t]
    \centering
    \includegraphics[width=1\linewidth]{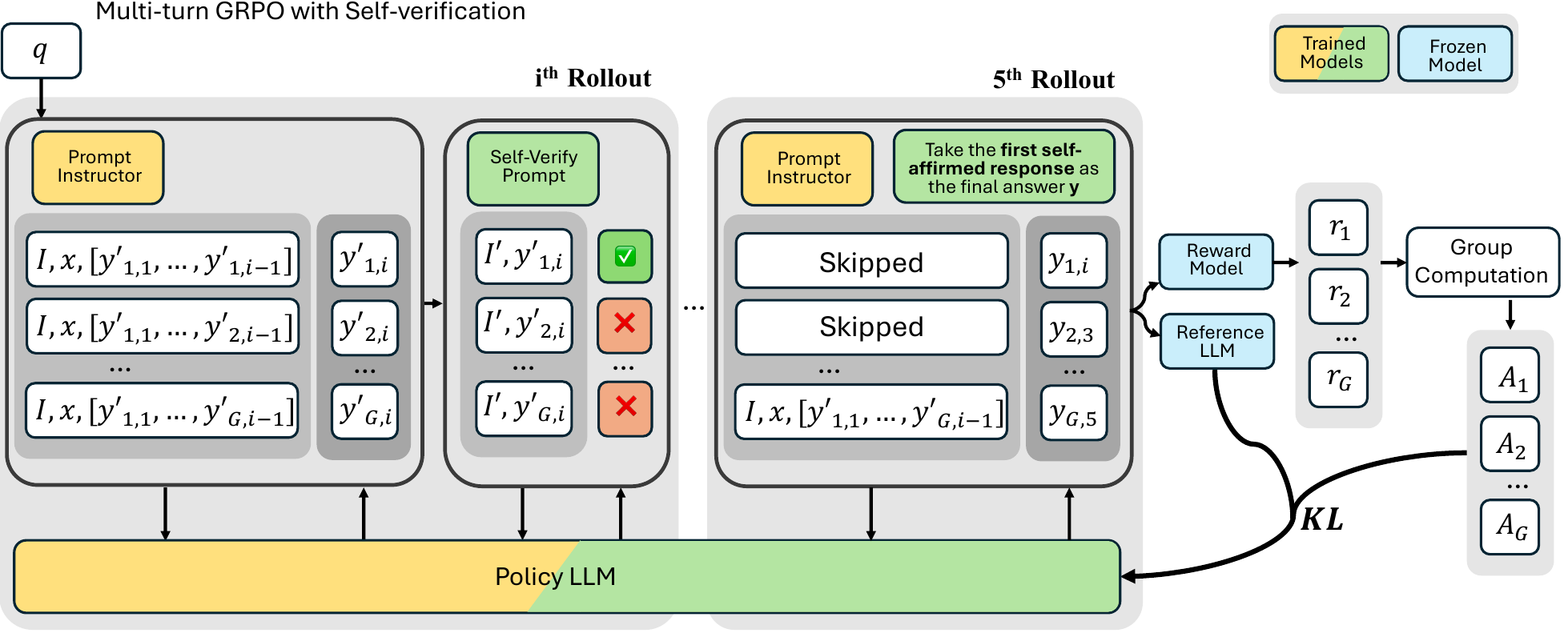}
    \caption{Multi-Modal GRPO w. Self Verification Training Pipeline.}
    \vspace{-1.5em}
    \label{fig:tooluse_grpo}
\end{figure*}

\noindent \textbf{Self-verification Protocol.}
\textbf{SVR-R1} adopts binary feedback for the verifier, restricting its output to only \textit{Yes} or \textit{No} via instructing it with the proper text prompts. This is because VLMs are generally known to be less capable than LLMs at providing detailed feedback with rationales, and prior work \citep{vlm-self-verify} has shown simple binary feedback to be the most effective in the VLM domain at minimizing the risk of hallucinations. \cref{fig:binary_verifier} in Appendix depicts the prompt instruction $x_v$ \textbf{SVR-R1} uses for the self-verifier, constraining it to only indicate whether the previous prediction is correct or not. Empirically, we find that existing VLMs reliably follow this instruction, and the sampled output from self-verification is consistently either ``YES'' or ``NO''. Formally, in the self verification response $y' \sim \pi_{\theta}(\cdot \mid I, x')$, (1) $\theta$ is the same copy of parameters as generator (described shortly), (2) $x'$ includes all of the multi-modal question, the response in previous self-generation turn, and the aforementioned instruction prompt for verification, and (3) $y' \in\{\mathrm{YES},\mathrm{NO}\}$. For example, in the verification round following the initial generation, input prompt can be denoted as $x' = x \oplus y_0 \oplus x_v$, where $\oplus$ is simple concatenation.


\noindent \textbf{Self-(Re)generation Protocol.} In the self-(re)generation step, if the obtained binary self-verification result $y' = \mathrm{No}$ and a user-specified maximum number of turns has not been reached, the model will be prompted (via appending a textual rethink trigger $x_r$ to the self-verifier's response, \cref{fig:self_regeneration} in Appendix) to generating a revised response in a new assistant turn, while keeping previous turns in mind. 

Formally, if the verifier disagrees at the $i_{th}$ step, the new re-generated answer is sampled from the same VLM policy $\pi_{\theta}$, with the input recursively constructed by appending the previous response and a rethink trigger to the input from the previous round: $y_{i+1} \sim \pi_{\theta}(\cdot \mid I, x_{i})$, where $x_{i} = x_{i-1} \oplus y_i \oplus x_v$. This recursive multi-turn process encourages the model to allocate additional reasoning tokens and regenerate its answer whenever it fails self-verification, effectively simulating a verifier-driven "re-thinking" process without relying on external knowledge. Finally, when either $y' = \mathrm{yes}$ or the turn limit has been reached, the finalized answer is used for accuracy measurement (during inference) or reward assignment (in RL rollouts). 
\textbf{\cref{fig:method} depicts an overview of this process:} the model performs repeated generation-verification cycles from the 1st to the $(i-1)$th step, with each rollout becoming part of the context for the next. Only the final answer, once verification succeeds or the turn limit is reached, interacts with the outcome reward judge. No process rewards are assigned to verification. 

\subsection{Multi-turn RL with Self-verification}
\label{sec:method_rl}
This section describes \textbf{SVR-R1}'s \textit{multi-modal and multi-turn} RL procedure illustrated in \cref{fig:tooluse_grpo}, which aims to optimize for the final reasoning response $y$ with VLM rollouts including self-verifications. 

\textbf{SVR-R1}'s training objective is to encourage the VLM policy $\pi_{\theta}$ to improve its reasoning ability with iterative self-verification, while penalizing large update step from the reference model $\pi_{\text{ref}}$. Formally, we maximize the following training objective:
\begin{equation}
\label{eq:rl-tooluse}
\max_{\pi_\theta}
\mathbb{E}_{[I, T] \sim \mathcal{D},\; y \sim \pi_{\theta}(\cdot \mid I, T; \pi_\theta)}
\bigl[ r_{\phi}(I, T, y) \bigr]
- \beta \,
\mathbb{D}_{\mathrm{KL}} \!\bigl[
\pi_{\theta}(\cdot \mid I, T; \pi_\theta)
\,\|\, 
\pi_{\mathrm{ref}}(\cdot \mid I, T; \pi_\theta)
\bigr]
\end{equation}

\noindent where $\pi_{\theta}$ is the trainable VLM policy, $\pi_{\text{ref}}$ is the frozen reference model, $r_{\phi}$ is the reward function, and $\beta > 0$ is the KL penalty coefficient. The input $[I, T]$ represents multimodal samples (image $I$ and text question query $T$) drawn from dataset $\mathcal{D}$. \textit{We explicitly include $\pi_\theta$ in the sampling formulation to indicate that, during VLM policy generation, the policy itself also serves as the verifier.}

\noindent \textbf{End-to-end Optimization Objective for Multi-turn Rollout}. Unlike prior RL fine-tuning approaches that optimize single-pass generations~\citep{ouyang2022traininglanguagemodelsfollow, grpo}, \textbf{SVR-R1} explicitly incorporates multi-turn self-verification into rollouts. Each rollout includes intermediate verifier binary feedback steps, where the model itself decides whether to rethink. Thus, $y \sim \pi_{\theta}(\cdot \mid I, T)$ can denote either a direct answer or an answer refined after multiple verification turns. Notably, \textbf{SVR-R1} does not directly optimize the intermediate verification responses $y'$; it instead only optimizes the final response $y$ via outcome-based reward---either the first response that receives a "YES" from the verifier or the last response generated after reaching the predefined maximum number of turns (\cref{sec:method_protocol}), while allowing the model to autonomously decide whether invoking additional verification steps improves reasoning, incentivizing accurate, confident final answers and the generation of self-verifiable reasoning traces.

\noindent \textbf{Loss Masking for Verification Tokens.}
In RL finetuning \citep{grpo,schulman2017proximal}, losses are computed over the entire rollout sequence. In \textbf{SVR-R1}, however, the rollout includes multiple turns of both self-generated tokens and self-verification tokens (Yes or No). Since our objective is to optimize the final generation response—and the trigger "verifier disagrees" is already manually incorporated into the generation—directly including verification tokens in the loss calculation can introduce unintended learning dynamics. (For example, optimizing both self-verification and generation simultaneously may lead to conflicting training objectives.) Instead, our focus is on improving end-to-end reasoning performance with self-verification in the loop. Inspired by the multi-turn RL framework Search-R1~\citep{jin2025searchr1trainingllmsreason}, we mask out the self-verification tokens during loss calculation to stabilize policy updates during training.

\noindent \textbf{GRPO with Self-verification.} Specifically, \textbf{SVR-R1}'s optimization for parameters $\theta$ builds on Group Relative Policy Optimization (GRPO)~\citep{grpo}, a stable and resource-efficient online policy-gradient algorithm (\cref{fig:tooluse_grpo}), without a learned value function approximation as in PPO~\citep{schulman2017proximal}. GRPO estimates baselines from a group of Monte-Carlo sampled rollouts, eliminating the need for a critic, reducing training overhead and costs and demonstrating strong empirical performance \cite{grpo}. \textit{This efficiency makes GRPO particularly well-suited \textbf{SVR-R1} complex, multi-turn setting where training costs may otherwise be prohibitive.} Concretely, for each input $[I, T]$, \textbf{SVR-R1} samples a group of responses including self-verification steps, 
$\{ y_i \}_{i=1}^{G} \sim \pi_{\text{old}}(\cdot \mid I, T; \pi_{\text{old}}),$
from the old policy. The current policy $\pi_{\theta}$ is then updated by maximizing the group-relative objective:  
\vspace{-0.5em}
\begin{equation}
\label{eq:grpo_condensed_fig}
\mathcal{L}_{\text{GRPO}}(\theta)=
\mathbb{E}\!\left[\frac{1}{G}\sum_{i=1}^{G}
\min\!\Big(
r_i(\theta)\widehat{A}(y_i),\,
\mathrm{clip}\!\left(r_i(\theta),1-\epsilon,1+\epsilon\right)\widehat{A}(y_i)
\Big)\right]
-\beta\,\mathrm{D}_{\mathrm{KL}}\!\left(\pi_\theta \,\|\, \pi_{\mathrm{old}}\right),
\end{equation}
\vspace{-0.5em}
where \[
r_i(\theta)=
\frac{\pi_{\theta}(y_i \mid I,T;\pi_\theta)}
     {\pi_{\mathrm{old}}(y_i \mid I,T;\pi_{\mathrm{old}})}.
\]

\noindent where $\widehat{A}(y_i)$ is the group-relative advantage for response $y_i$, computed by normalizing outcome rewards within the group, and positive $\epsilon$ is the clip threshold. This formulation encourages exploration of diverse reasoning strategies, while maintaining stability and ensuring that the policy improves relative to its peers within the sampled group.

\noindent \textbf{Reward Design}.
\textbf{SVR-R1} utilizes an \textit{outcome-based binary reward} without format rewards. For complex, semi-open-form visual question answering tasks, such as the visual table and chart reasoning dataset in ReFocus (\cref{sec:exp_dataset}), we follow prior work~\citep{fu2025refocus} and adopt a large language model judge to compare the ground truth with the final prediction, with detailed prompts provided in the supplementary materials. For verifiable reasoning tasks, such as geometry math or multiple-choice questions, we use a rule-based judge.

\section{Experiment}
In this section, we empirically evaluate the effectiveness of SVR in multi-modal reasoning, present the findings and analyze the key factors contributing to SVR-R1's success.

\subsection{Experiment Setup}
\label{sec:exp_dataset}
\noindent \textbf{Dataset.} We use three dataset splits containing challenging table and chart reasoning tasks for proper assessment and benchmarking of performance. First two splits are prepared following the data pre-processing pipeline in ReFocus~\citep{fu2025refocus}:\begin{enumerate}
    \item \textbf{ChartQA Split}~\citep{masry2022chartqa} contains 826 test QA pairs split into 444 horizontal and 382 vertical bar chart questions, each requiring logical comparisons and visual reasoning over chart structures. We use the official ReFocus training set comprising 14,344 examples selected from the ChartQA training split (out of 15,059 available) for training.
    \item \textbf{TableVQA Split}~\citep{kim2024tablevqa} contains 1,250 table-based questions incorporating the VWTQ, VWTQ-syn, and VTabFact splits. We allocate 70\% of questions for training and 30\% for testing.
    \item \textbf{ThinkLite-VL-70K}  from recent multimodal reasoning work~\citep{wang2025sotalessmctsguidedsample}, which includes general multimodal reasoning tasks for training.
\end{enumerate}
Full dataset details about \textbf{ThinkLite-VL-70K} are in the supplementary materials.

\noindent \textbf{Implementation.}
We build \textbf{SVR-R1} upon a state-of-the-art open-source VLM  Qwen-VL 2.5 \citep{qwenvl25} at the 3B and 7B scales. We implement \textbf{SVR-R1} with GRPO (\cref{sec:method_rl}) in the open-source VeRL framework \citep{sheng2024hybridflow} for multi-modal and multi-turn RL fine-tuning. Additional implementation details are provided in \cref{sec:details} of the Appendix.

\noindent \textbf{Methods.} 
We compare \textbf{SVR-R1} against the following notable methods:\begin{enumerate}
    \item \textbf{Qwen2.5-VL} \citep{qwenvl25}: The baseline model which \textbf{SVR-R1} builds upon; we use it without \textbf{SVR-R1}'s training.
    
    \item \colorbox{blue!10}{\textbf{GPT-4o}}~\citep{openai2024openaio1card}: a commercial VLM which we access via API. We use the 2024-08-06 checkpoint \citep{openai_gpt4o_2024} for maximal reproducibility.

    \item \colorbox{orange!10}{\textbf{R1-VL}}~\citep{zhang2025r1vllearningreasonmultimodal}: a recently open-sourced reasoning VLM RL-trained for general reasoning tasks in academic. 

    \item \colorbox{gray!08}{\textbf{Weaker Baselines:}} As included in the original ReFocus paper~\citep{fu2025refocus}, these comprise classic models such as Llava-next~\citep{liu2024llavanext}, Phi-3~\citep{abdin2024phi3technicalreporthighly}, Gemini 1.5~\citep{geminiteam2025geminifamilyhighlycapable}, and VisProg~\citep{visprog}.
\end{enumerate}
Additionally, we use an ablated version of Qwen-VL 2.5, \colorbox{red!25}{\textit{Qwen-RL}}, which we train with standard GRPO with exact the same data and hyperparameter setting, but do not perform any self-verification rounds in the rollout, to study the impact of self-verification on reasoning performance. 

\noindent \textbf{Inference Setup.} To demonstrate how \textbf{SVR-R1} internalizes self-verification advantages, we evaluate models under two distinct settings:
\begin{enumerate}
\item \textbf{Inference Pure Run (PR)}: Direct inference run without any self-verification step.
\item \textbf{Inference with Final Verification (FV)}: Inference run that includes self-verification and rethinking steps, following the same protocol used during training rollouts. This process continues until either an early \textsc{Yes} is obtained or a maximum of $MAX=3$ iterations is reached.
\end{enumerate}
\noindent \textbf{Reward Judge.} We employ the state-of-the-art open-source large language model, gpt-oss-120b~\citep{openai2025gptoss120bgptoss20bmodel}, to evaluate model predictions against ground-truth answers. The reward is binary: 1 for correct predictions and 0 for incorrect ones. For the ThinkLite experiments, we follow the original work \citep{wang2025sotalessmctsguidedsample} to use a rule-based binary reward to match the answer.

\begin{table}[t]
\centering
\setlength{\tabcolsep}{2pt}
\scriptsize
\resizebox{\columnwidth}{!}{%
\begin{tabular}{l|cccccccccccc}
\toprule[1pt]
\multirow{2}{*}{Qwen2.5-VL} 
& \multicolumn{6}{c}{\textbf{3B}} 
& \multicolumn{6}{c}{\textbf{7B}} \\
\cmidrule(lr){2-7} \cmidrule(lr){8-13}
& PR & FV & \cellcolor{red!25}RL-PR & \cellcolor{red!25}RL-FV & \cellcolor{green!25}SVR-PR & \cellcolor{green!25}SVR-FV
& PR & FV & \cellcolor{red!25}RL-PR & \cellcolor{red!25}RL-FV & \cellcolor{green!25}SVR-PR & \cellcolor{green!25}SVR-FV \\
\hline
ChartQA Split  & 63.3 & 64.9 & \cellcolor{red!10}{80.5} & \cellcolor{red!10}{80.9} & \cellcolor{green!10}{83.3} & \cellcolor{green!10}{83.3} & 76.0 & 76.9 & \cellcolor{red!10}{80.4} & \cellcolor{red!10}{81.1} & \cellcolor{green!10}{82.9} & \cellcolor{green!10}{82.9} \\
TableVQA Split & 56.4 & 59.0 & \cellcolor{red!10}{68.3} & \cellcolor{red!10}{68.7} & \cellcolor{green!10}{72.4} & \cellcolor{green!10}{72.9} & 67.8 & 68.6 & \cellcolor{red!10}{78.7} & \cellcolor{red!10}{78.5} & \cellcolor{green!10}{80.3} & \cellcolor{green!10}{80.6} \\
\bottomrule[1.2pt]
\end{tabular}%
}
\caption{Main results (accuracy in \%) of raw Qwen2.5-VL, \colorbox{green!25}{SVR-R1}, and \colorbox{red!25}{Qwen-RL}. \textit{PR: Pure Run; FV: with Final Verification.}}
\label{tab:main}
\end{table}
\subsection{Empirical Findings}
We conduct controlled experiments using \textbf{identical datasets and hyperparameters }for both standard GRPO and SVR-R1, and find that \textbf{SVR-R1 significantly boosts multimodal reasoning with same amount of data}. Table~\ref{tab:main} and Table~\ref{tab:second} present our main results, highlighting the effectiveness of \textbf{SVR-R1} for fine-tuning VLM on challenging visual table and chart reasoning tasks. At both the 3B and 7B model scales, SVR-R1 consistently outperforms the baseline GRPO methods by a substantial margin, even though GRPO already demonstrates strong capabilities in fine-tuning models for these tasks. Similar patterns exist in the general reasoning dataset with larger scale, as detailed in Table~\ref{tab:third}, models trained on ThinkLite data exhibit improved performance on more general reasoning tasks.

\newcolumntype{C}[1]{>{\centering\arraybackslash}m{#1}}

\newcommand{\modelhead}[1]{%
  \makebox[0pt][l]{%
    \hspace{0.2em}%
    \raisebox{0.15em}[0pt][0pt]{%
      \rotatebox[origin=lb]{62}{\fontsize{5pt}{5.5pt}\selectfont\strut #1}%
    }%
  }%
}

\begin{wrapfigure}{r}{0.5\columnwidth}
  \centering
  \vspace{-1.8\baselineskip}
  \includegraphics[width=0.4\columnwidth]{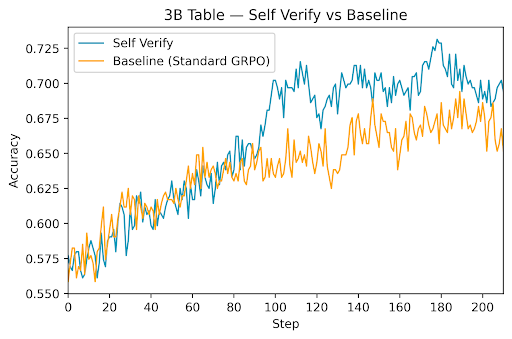}
  \vspace{-0.8em}
  \caption{SVR-R1 surpasses standard GRPO on table tasks.}
  \label{fig:table-plot}
   \vspace{-1.5\baselineskip}
\end{wrapfigure}

\begin{table}[t]
\centering
\setlength{\tabcolsep}{1.2pt}
\scriptsize
\renewcommand{\arraystretch}{1.05}
\resizebox{\columnwidth}{!}{%
\begin{tabular}{l|*{12}{C{2.55em}}}
\toprule
\multicolumn{1}{c|}{\rule{0pt}{3.7em}}
& \modelhead{R1VL-2B-PR}
& \modelhead{R1VL-2B-FV}
& \modelhead{R1VL-7B-PR}
& \modelhead{R1VL-7B-FV}
& \modelhead{GPT4o-PR}
& \modelhead{GPT4o-FV}
& \modelhead{LLaVA-34B}
& \modelhead{Phi-3V}
& \modelhead{Gem-Pro1.5}
& \modelhead{VisProg}
& \modelhead{SVR-3B}
& \modelhead{SVR-7B}
\\[-0.2em]
\midrule
Chart
& \colorbox{orange!10}{53.6}
& \colorbox{orange!10}{52.7}
& \colorbox{orange!10}{73.2}
& \colorbox{orange!10}{73.1}
& \colorbox{blue!10}{77.8}
& \colorbox{blue!10}{79.3}
& \colorbox{gray!08}{43.7}
& \colorbox{gray!08}{52.3}
& \colorbox{gray!08}{46.9}
& \colorbox{gray!08}{59.6}
& \colorbox{green!25}{83.3}
& \colorbox{green!25}{82.9} \\
Table
& \colorbox{orange!10}{37.8}
& \colorbox{orange!10}{34.8}
& \colorbox{orange!10}{58.2}
& \colorbox{orange!10}{57.4}
& \colorbox{blue!10}{76.9}
& \colorbox{blue!10}{76.6}
& \colorbox{gray!08}{18.4}
& \colorbox{gray!08}{63.4}
& \colorbox{gray!08}{61.3}
& \colorbox{gray!08}{69.2}
& \colorbox{green!25}{72.9}
& \colorbox{green!25}{80.6} \\
Extra SFT
& \cmark
& \cmark
& \cmark
& \cmark
& \xmark/\cmark
& \xmark/\cmark
& \cmark
& \xmark/\cmark
& \xmark/\cmark
& \xmark
& \xmark
& \xmark \\
\bottomrule
\end{tabular}%
}
\caption{Ours vs. other models. Colored groups: \colorbox{green!25}{SVR-R1}, \colorbox{orange!10}{R1-VL}, \colorbox{blue!10}{GPT-4o}, and \colorbox{gray!08}{weaker baseline models} in Chart and Table Reasoning Tasks.}
\label{tab:second}
\end{table}
\begin{wraptable}{r}{0.5\columnwidth}
\vspace{-0.8em}
\centering
\setlength{\tabcolsep}{3.5pt}
\scriptsize
\renewcommand{\arraystretch}{1.05}
\begin{tabular}{l|cccc}
\toprule
7B ThinkLite & MathVista & MathVision & MMStar & AI2D \\
\midrule
\rowcolor{red!10} RL-BEST & 70.8 & 17.4 & 48.7 & \textbf{81.5} \\
\rowcolor{green!25} SVR-R1 & \textbf{71.6} & \textbf{19.1} & \textbf{49.3} & 81.4 \\
\bottomrule
\end{tabular}
\vspace{-0.4em}
\caption{\colorbox{green!25}{SVR-R1} outperforms standard GRPO on general reasoning when trained in ThinkLite under controlled implementations.}
\label{tab:third}
\vspace{-1.0em}
\end{wraptable}
To further illustrate the training dynamics, we provide train-test plots for both our method and standard GRPO in the thumbnail Figure~\ref{fig:thumb-acc-1col} for chart reasoning, and detailed results for table tasks in Figure~\ref{fig:table-plot}. These figures clearly show that incorporating self-verification into RL training both accelerates learning under the same data budget and improves the final saturated performance. Given the difficulty of curating high-quality relevant multimodal RL datasets, effective self-verification during RL training is especially valuable, enabling stronger post-training results on the target tasks.

\noindent \textbf{Robustness of Gain.} Comparing the table and chart tasks, we observe that chart reasoning benefits from a larger dataset and a more stable training curve, while table reasoning is limited by a smaller training set (only several hundred samples), resulting in a comparatively noisier curve in Figure~\ref{fig:table-plot}. Nevertheless, our method demonstrates robust performance gains even under less ideal data conditions for table tasks.

\noindent \textbf{Decreased Verification Turns and Increased Confidence.} Throughout training, we observe that both validation and training verification turns gradually decrease, as the models become increasingly confident in their initial answers and tend to affirm their responses: In Figures~\ref{fig:thumb-acc-1col} and \ref{fig:table-turns}, these curves converge to approximately 2 - one generation step followed by a \textsc{Yes} from the self-verifier round. For the chart task, SVR-R1 achieves nearly identical accuracy in the pure-run setting (without final verification) and the final-verification setting during the later stages of training, as reported in Table~\ref{tab:main}. This indicates that VLMs are able to internalize the verification and generation gap during SVR-R1 training, providing self-assured answers when correct—even without directly optimizing for self-verification capabilities in our RL training process. Intuitively, this is a desirable outcome, as it indicates that the VLM can produce correct answers while recognizing their correctness in future tasks. And it brings benefits in the inference efficiency if we do not need to go over long chain of self correction.

\begin{figure}[h]
  \includegraphics[width=0.48\linewidth]{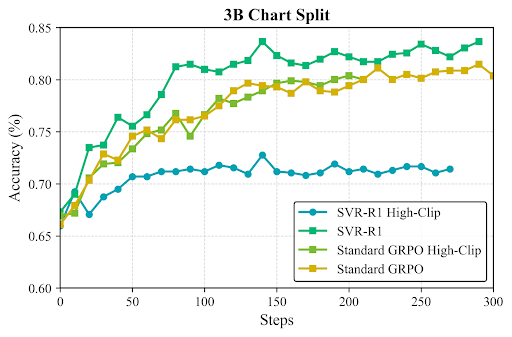}
  \hfill
  \includegraphics[width=0.48\linewidth]{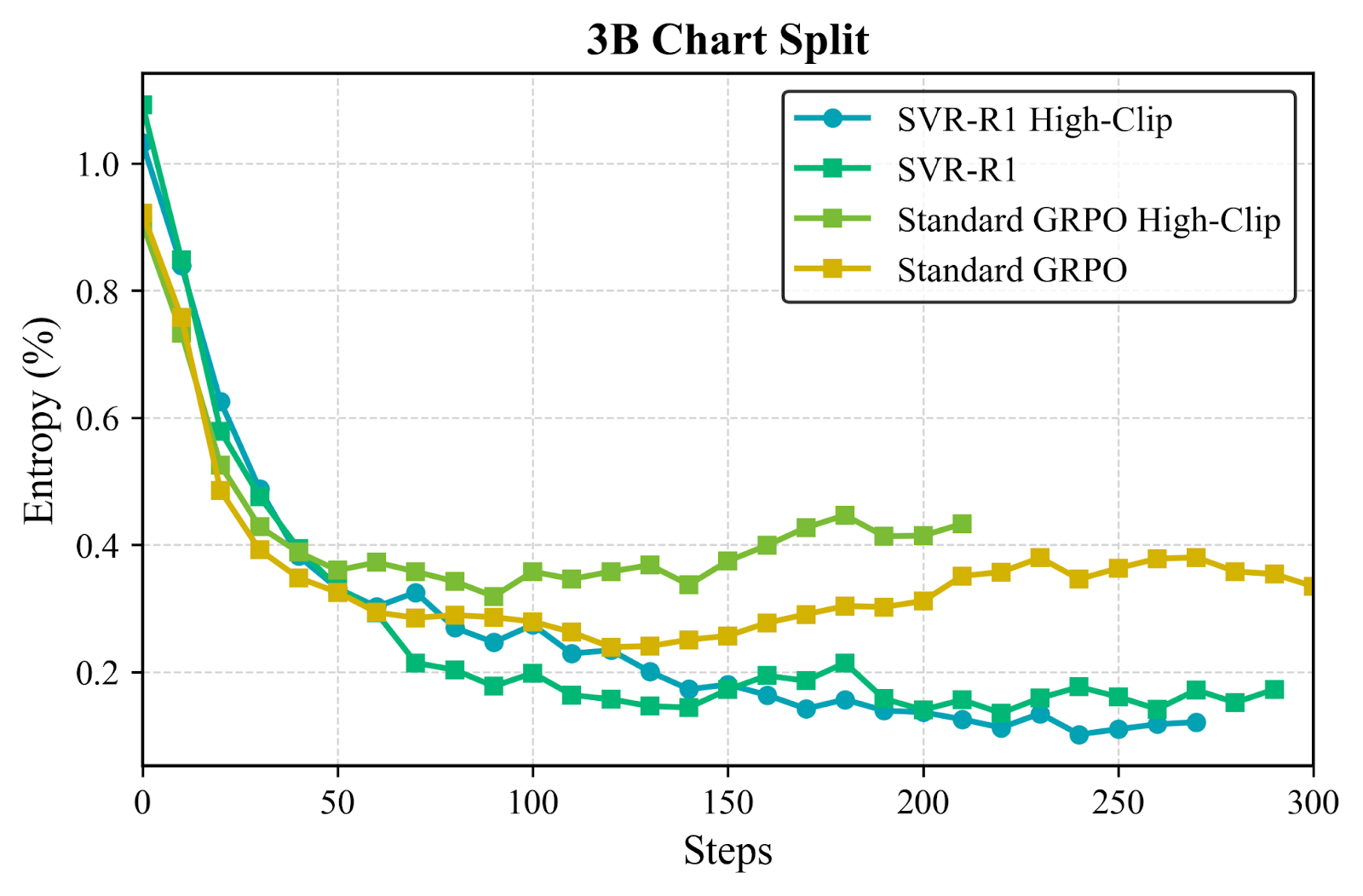}
  \vspace{-0.8em}
\caption{SVR-R1 outperforms high-entropy baselines. High-entropy variants are trained with $\epsilon_h$ = 2.8, compared to the standard setting of $\epsilon_h$ = 2.0. \textbf{Accuracy, Entropy} vs. \textbf{Steps.}}
  \label{fig:chart-entropy}
\end{figure}

\noindent \textbf{Lower Entropy.} We observe that incorporating self-verification into RL training consistently results in a model with relatively lower entropy for a given task compared to standard GRPO. While this reduction reflects increased confidence, however, excessively trading exploration for accuracy is not always desirable in reasoning tasks that require exloration such as math \citep{yu2025dapoopensourcellmreinforcement}. To ensure a fair comparison, we employ an established entropy-controlled technique from DAPO \citep{yu2025dapoopensourcellmreinforcement}, applying a higher clip-high threshold $\epsilon_h$ for RL training, to assess whether the lower entropy observed in our approach negatively impacts performance on chart tasks in 3B model training. Specifically, we investigate whether increasing entropy over our solutions or standard GRPO can yield performance gains in our task settings. The results, presented in Figure~\ref{fig:chart-entropy}, reveal two key findings: (1) higher entropy does not lead to improved results in our tasks, even in the standard GRPO setting, and (2) self-verification does not benefit from high clip techniques which preserve higher entropy. These experiments demonstrate that SVR-R1 does not "over-sacrifice" entropy for accuracy, and maintains robust performance without compromising necessary exploration.

\noindent \textbf{Eliciting Self-reflection via Rethinking Trigger.} We present a real validation example in the Appendix Figure~\ref{fig:cat-breed-verification}, demonstrating that SVR-R1 explicitly prompts the model to engage in self-reflection, moving beyond simple rejection and regeneration. When the self-verifier returns \textsc{no}, we prompt the model to reflect directly (see Figure~\ref{fig:self_regeneration}), successfully correcting its wrong answer in this cat breed identification example. Notably, even though the verifier does not successfully affirm the correct answer in the second round, we still observe benefits from rethinking in the third round: the model still arrives at the correct answer "calico" and adopts more conservative phrasing, such as \textbf{"also acknowledge the possibility"}. Another interesting finding is that after the rethinking phrase \textbf{"given the verifier's disagreement,"} the model explores diverse reasoning paths, such as physical features or visible characteristics, with increasing confidence across rounds.

\vspace{-1.2em}

\vspace{-0.8em}
\noindent \textbf{VLMs in the context of Inference-Time Scaling.} We attribute part of \textbf{SVR-R1}'s performance gains to its effective use of existing VLM capabilities during scaled inference and rollout, even without requiring highly sophisticated self-verification mechanism. As self-verification capabilities of VLMs continue to improve, important research questions would emerge: How can we optimize VLM policy with inference-time scaling directly through RL? \textbf{SVR-R1} offers an early and promising step forward by demonstrating that integrating self-verification into both training and inference-time reasoning yields substantial benefits.

\noindent \textbf{Squeeze What Models Can Answer with Effort.} \textbf{SVR-R1}'s performance gain can be attributed to it finalizing high-quality answers to medium difficulty questions in the datasets via multiple rollout rounds. In contrast, repeatedly rethinking on questions that are \textit{too difficult} contributes little to training, as the correct answer may remain unreachable even with unlimited self-verification turns. We further validate this by training with self-verification only on a dataset of difficult questions \citep{wang2025sotalessmctsguidedsample}, where we observe no improvement in reasoning performance while the number of verification turns grows dramatically during RL training.
This performance gain brought by focusing efforts in solving medium-difficulty questions has been observed in prior works such as \citep{gao2025promptcurriculumlearningefficient} in the language domain, have demonstrated benefits by creating training curriculum with batches of intermediate difficulty tailored to current model weights in current training step. 

\section{Conclusion}
We introduced \textbf{Self-Verified Reasoner (SVR-R1)}, which integrates self-verification rounds into GRPO training to bootstrap VLM's reasoning capabilities. \textbf{SVR-R1} significantly outperforms standard GRPO on multi-modal reasoning benchmarks with exactly the same data and hyperparameter setup. Notably, models gradually perform fewer verification rounds during training while maintaining improved accuracy, suggesting they learn to close the generation-verification gap by producing initial answers that pass self-verification. This work demonstrates that models can effectively leverage their inherent verification capabilities for self-improvement within the RL training loop, advancing multimodal reasoning.

\bibliography{colm2026_conference}
\bibliographystyle{colm2026_conference}

\appendix
\section{Appendix}
\subsection{Qualitative Example}
\begin{figure}[htbp]
    \centering
    \begin{tcolorbox}[
        halign=flush left,
        breakable,
        colback=teal!5!white,
        colframe=teal!75!black,
        title={\footnotesize\textbf{Cat Breed Identification with Self-reflection}},
        fonttitle=\bfseries,
        fontupper=\ttfamily,
        width=\columnwidth,
        enhanced jigsaw,
        boxrule=0.5pt,
        left=1pt, right=1pt, top=1pt, bottom=1pt
    ]
    \scriptsize
    \linespread{1.0}\selectfont
    \setlength{\parskip}{0pt}
    \begin{minipage}[t]{0.62\linewidth}
    {\color{orange}user}
    {\color{blue}Which breed of cat is this?}\\
    {\color{orange}assistant}\\
    {\color{purple}<think>}
    {\color{black}The cat in the picture appears to be a domestic shorthair. This is a common breed that can have a variety of coat patterns and colors. }\\
    \end{minipage}%
    \hfill
    \begin{minipage}[t]{0.35\linewidth}
    \vspace{0pt}
    \includegraphics[width=\linewidth]{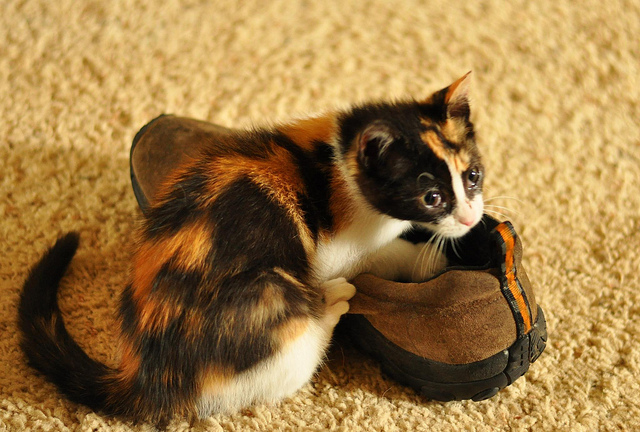}
    \end{minipage}
    {\color{black}The specific markings and coloration suggest it could be a calico or a tortoiseshell, which are common patterns within the domestic shorthair category. However, without more detailed information or a clearer view of the cat's physical features, it's not possible to determine the exact breed with certainty.}
    {\color{purple}</think>}
    {\color{brown}\textbackslash boxed\{tabby\}}\\
    
    {\color{orange}user}\\
    {\color{blue} A verifier disagrees with your response. Think again.}\\
    
    {\color{orange}assistant}
    {\color{purple}<think>}
    {\color{black}\textbf{Given the verifier's disagreement, I will re-evaluate }the cat's breed based on the visible characteristics. The cat in the picture has a mix of white and black fur with some brown patches, which is typical of a calico or tortoiseshell pattern. However, the specific markings and coloration can vary widely within these patterns, and without more detailed information or a clearer view of the cat's physical features, it's not possible to determine the exact breed with certainty.}
    {\color{black}The cat's ears are pointed, which is a common feature in many breeds, including domestic shorthairs. The overall appearance, including the coat pattern, suggests it could be a domestic shorthair, which is a common and diverse breed that can have a variety of coat patterns and colors.}  
    {\color{black}\textbf{Given the verifier's disagreement}, I will consider the possibility that the cat could be a domestic shorthair with a calico or tortoiseshell pattern.}
    {\color{purple}</think>}
    {\color{brown}\textbackslash boxed\{calico\}}\\
    
    {\color{orange}user}\\
    {\color{blue}verifier disagrees with your response. Think again.}\\
    
    {\color{orange}assistant}
    {\color{purple}<think>}
    {\color{black}Given the verifier's disagreement, I will re-evaluate the cat's breed based on the visible characteristics. [...] \textbf{Given the verifier's disagreement, I will consider} the possibility that the cat could be a domestic shorthair with a calico or tortoiseshell pattern, \textbf{but I will also acknowledge the possibility} that it could be another breed with similar markings.}
    {\color{purple}</think>}
    {\color{brown}\textbackslash boxed\{calico\}}\\
    
    {\color{red}Ground Truth:} {\color{brown}calico} {\color{red}| Score:} {\color{brown}1.0}
    \end{tcolorbox}
    \vspace{-10pt}
    \caption{Self-reflection of trained model in an example from \textbf{ThinkLite-70K} \citep{wang2025sotalessmctsguidedsample}.}
    \vspace{-7pt}
    \label{fig:cat-breed-verification}
\end{figure}

We list one full qualitative example in \cref{fig:cat-breed-verification}.
\subsection{Full Prompt for VLM Rollout and Inference}
\begin{figure}[htbp]
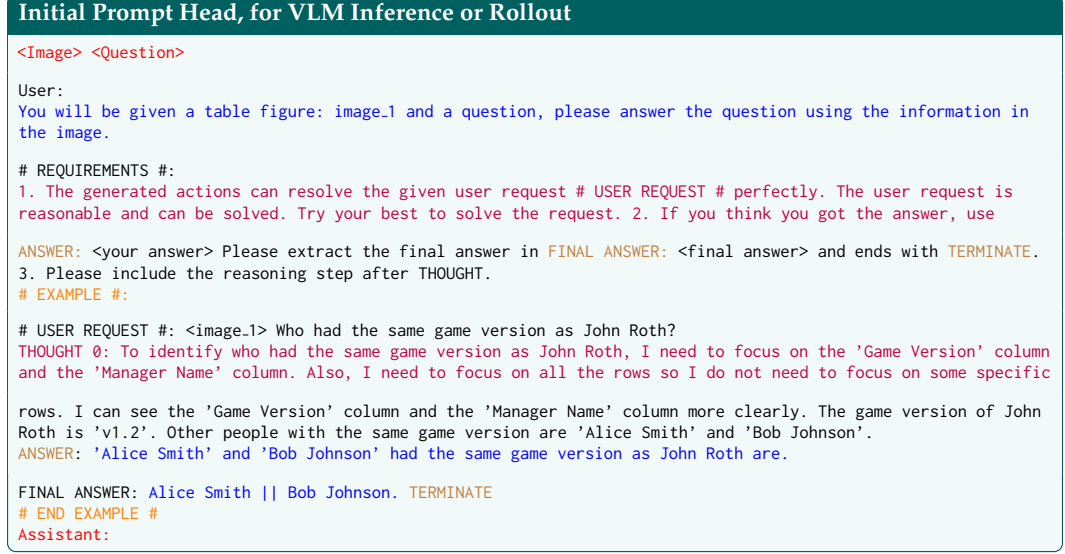

    \centering
    \begin{tcolorbox}[
        halign=flush left,
        breakable,
        colback=teal!5!white,
        colframe=teal!75!black,
        title={\footnotesize\textbf{Initial Prompt Head, for VLM Inference or Rollout}},
        fonttitle=\bfseries,
        fontupper=\ttfamily,
        width=\columnwidth, 
        enhanced jigsaw,
        boxrule=0.5pt,
        left=1pt, right=1pt, top=1pt, bottom=1pt
    ]
    \scriptsize
    {\color{red}<Image>} {\color{red}<Question>}

    {\color{red}User:}  

    {\color{blue}You will be given a table figure: image\_1 and a question, please answer the question using the information in the image.}  

    {\color{orange}\# REQUIREMENTS \#:}  

    {\color{purple}1. The generated actions can resolve the given user request \# USER REQUEST \# perfectly. The user request is reasonable and can be solved. Try your best to solve the request.}
    {\color{purple}2. If you think you got the answer, use {\color{brown}ANSWER:} <your answer> Please extract the final answer in {\color{brown}FINAL ANSWER:} <final answer> and ends with {\color{brown}TERMINATE}. 3. Please include the reasoning step after THOUGHT}. 
    
    {\color{orange}\# EXAMPLE \#:}  

    {\color{blue}\# USER REQUEST \#: <image\_1> Who had the same game version as John Roth?}

    {\color{purple}THOUGHT 0: To identify who had the same game version as John Roth, I need to focus on the 'Game Version' column and the 'Manager Name' column. Also, I need to focus on all the rows so I do not need to focus on some specific rows. I can see the 'Game Version' column and the 'Manager Name' column more clearly. The game version of John Roth is 'v1.2'. Other people with the same game version are 'Alice Smith' and 'Bob Johnson'.
} 
    
    {\color{brown}ANSWER}: {\color{blue}'Alice Smith' and 'Bob Johnson' had the same game version as John Roth are.}

    {\color{brown}FINAL ANSWER:} {\color{blue}Alice Smith || Bob Johnson. {\color{brown}TERMINATE}}  

    {\color{orange}\# END EXAMPLE \#}

    {\color{red}Assistant:}  
    \end{tcolorbox}
    \vspace{-10pt}
    \caption{Full Prompt for VLM Inference or Rollout, with an In-context Example and Clear Requirements.}
    \vspace{-10pt}
    \label{fig:tcolorbox-vlm-head}
\end{figure}

\noindent We construct the initial prompt header (\cref{fig:tcolorbox-vlm-head}) for \textbf{SVR-R1} to include carefully designed requirements, explicit formatting instructions, and a one-shot example with detailed reasoning steps. This structure is intended to enhance multimodal reasoning with chain of thoughts \citep{CoT}, drawing inspiration from previous work ReFocus \citep{fu2025refocus}.
\begin{figure}[h]
\centering
\begin{tcolorbox}[
    halign=flush left,
    breakable,
    colback=teal!5!white,
    colframe=teal!75!black,
    title={\footnotesize\textbf{Self-Verifier Turn}},
    fonttitle=\bfseries,
    fontupper=\ttfamily,
    width=\columnwidth,
    enhanced jigsaw,
    boxrule=0.5pt,
    left=1pt, right=1pt, top=1pt, bottom=1pt
]\scriptsize
\textcolor{red}{<Image>} \textcolor{red}{<Question>} 
\textcolor{red}{<Initial Prompt Head>} 

\textcolor{red}{<Previous Responses in Assistant>}

\textcolor{red}{User:} "Evaluate whether your response is correct. If yes, output <YES>, if no, output <NO>. Do not make additional statements."

\textcolor{red}{Assistant:} 
\end{tcolorbox}
\vspace{-10pt}
\caption{Self-Verifier Turn, with Previous Queries and Response.}
\vspace{-10pt}
\label{fig:binary_verifier}
\end{figure}

\begin{figure}[h]
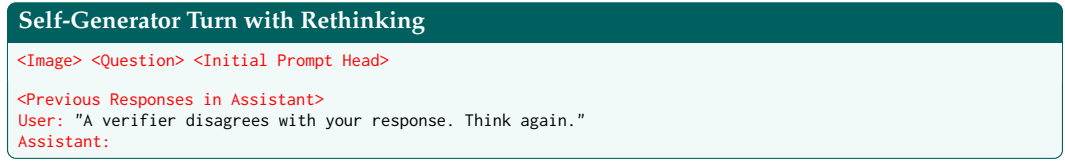

    \centering
    \begin{tcolorbox}[
        halign=flush left,
        breakable,
        colback=teal!5!white,
        colframe=teal!75!black,
        title={\footnotesize\textbf{Self-Generator Turn with Rethinking}},
        fonttitle=\bfseries,
        fontupper=\ttfamily,
        width=\columnwidth,
        enhanced jigsaw,
        boxrule=0.5pt,
        left=1pt, right=1pt, top=1pt, bottom=1pt
    ]\scriptsize
    \textcolor{red}{<Image>} \textcolor{red}{<Question>} \textcolor{red}{<Initial Prompt Head>}
    
    \textcolor{red}{<Previous Responses in Assistant>}
    
    \textcolor{red}{User:} "A verifier disagrees with your response. Think again."
    
    \textcolor{red}{Assistant:} 
    \end{tcolorbox}
    \vspace{-10pt}
    \caption{Self-Generator Turn with the Rethinking Trigger.}
    \vspace{-10pt}
    \label{fig:self_regeneration}
\end{figure}

\subsection{Full Prompt for Reward Judge}
\begin{figure}[htbp]
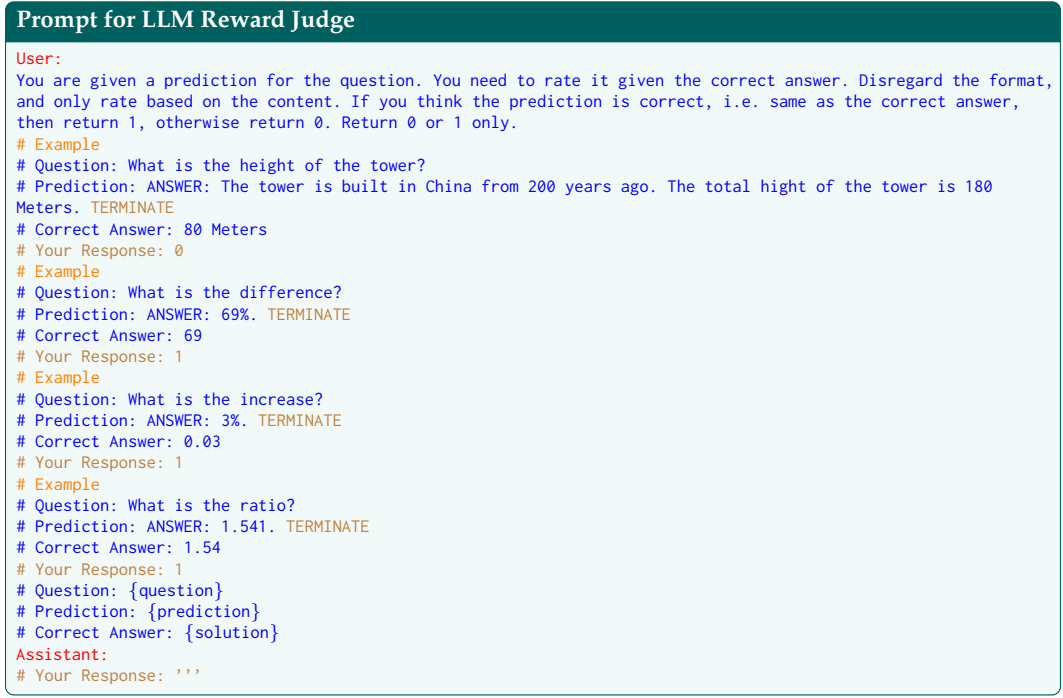

    \centering
    \begin{tcolorbox}[
        halign=flush left,
        breakable,
        colback=teal!5!white,
        colframe=teal!75!black,
        title={\footnotesize\textbf{Prompt for LLM Reward Judge}},
        fonttitle=\bfseries,
        fontupper=\ttfamily,
        width=\columnwidth, 
        enhanced jigsaw,
        boxrule=0.5pt,
        left=1pt, right=1pt, top=1pt, bottom=1pt
    ]
    \scriptsize
    {\color{red}User:}  
    
    {\color{blue}You are given a prediction for the question. You need to rate it given the correct answer. Disregard the format, and only rate based on the content. If you think the prediction is correct, i.e.\ same as the correct answer, then return 1, otherwise return 0. Return 0 or 1 only.}

    {\color{orange}\# Example}
    
    {\color{blue}\# Question: What is the height of the tower?}
    
    {\color{blue}\# Prediction: ANSWER: The tower is built in China from 200 years ago. The total hight of the tower is 180 Meters. {\color{brown}TERMINATE}}

    {\color{blue}\# Correct Answer: 80 Meters}
    
    {\color{brown}\# Your Response: 0}
    
    {\color{orange}\# Example}
    
    {\color{blue}\# Question: What is the difference?}
    
    {\color{blue}\# Prediction: ANSWER: 69\%. {\color{brown}TERMINATE}}
    
    {\color{blue}\# Correct Answer: 69}
    
    {\color{brown}\# Your Response: 1}

    {\color{orange}\# Example}
  
    {\color{blue}\# Question: What is the increase?}
    
    {\color{blue}\# Prediction: ANSWER: 3\%. {\color{brown}TERMINATE}}
    
    {\color{blue}\# Correct Answer: 0.03}
    
    {\color{brown}\# Your Response: 1}

    {\color{orange}\# Example}
    
    {\color{blue}\# Question: What is the ratio?}
    
    {\color{blue}\# Prediction: ANSWER: 1.541. {\color{brown}TERMINATE}}
    
    {\color{blue}\# Correct Answer: 1.54}
    
    {\color{brown}\# Your Response: 1}

    {\color{blue}\# Question: \{question\}}
    
    {\color{blue}\# Prediction: \{prediction\}}
    
    {\color{blue}\# Correct Answer: \{solution\}}
    
    {\color{red}Assistant:}
    
    {\color{brown}\# Your Response: '''}

    \end{tcolorbox}

    \vspace{-10pt}
    \caption{Prompt for LLM Judge in Semi-open Questions.}
    \vspace{-10pt}
    \label{fig:tcolorbox-pred-rating}
\end{figure}

\begin{figure}[htbp]
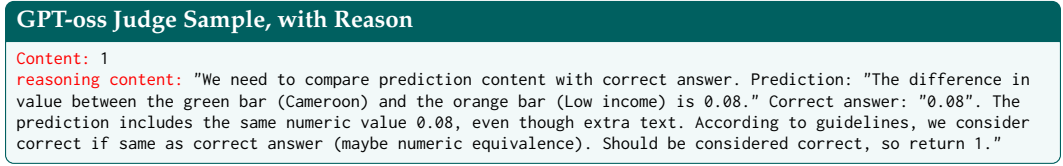

    \centering
    \begin{tcolorbox}[
        halign=flush left,
        breakable,
        colback=teal!5!white,
        colframe=teal!75!black,
        title={\footnotesize\textbf{GPT-oss Judge Sample, with Reason}},
        fonttitle=\bfseries,
        fontupper=\ttfamily,
        width=\columnwidth, 
        enhanced jigsaw,
        boxrule=0.5pt,
        left=1pt, right=1pt, top=1pt, bottom=1pt
    ]\scriptsize
    
    \textcolor{red}{Content:} 1
    
    \textcolor{red}{reasoning content:} "We need to compare prediction content with correct answer. Prediction: "The difference in value between the green bar (Cameroon) and the orange bar (Low income) is 0.08." Correct answer: "0.08". The prediction includes the same numeric value 0.08, even though extra text. According to guidelines, we consider correct if same as correct answer (maybe numeric equivalence). Should be considered correct, so return 1."
    \end{tcolorbox}
    \vspace{-10pt}
    \caption{GPT‑OSS provides high‑quality judgments, accompanied by step‑by‑step reasoning.}
    \label{fig:sample_judge}
\end{figure}

\textbf{SVR-R1} expects an \textit{outcome-based binary reward}. For complex, semi–open-form visual question answering tasks, such as the visual table and chart reasoning dataset in ReFocus, we follow prior work~\citep{fu2025refocus} and employ a LLM judge to directly compare the final prediction with the ground truth (\cref{fig:tcolorbox-pred-rating}). For this evaluation, we instruct the LLM to output a binary decision based solely on the prediction and the ground-truth answer. We use \textit{gpt-oss-120b}~\citep{openai2025gptoss120bgptoss20bmodel}, which performs well on this matching task with reasoning traces (\cref{fig:sample_judge}), and runs efficiently on a single 80~GB GPU using vLLM as the serving engine. Importantly, we do not introduce any additional information via the judge: the model is not asked to answer the question; it only determines whether the prediction matches the ground truth. To our knowledge, integrating an LLM-judge reward into RL training for VLMs is novel in this domain; prior work ~\citep{zhou2025r1zerosahamomentvisual, chen2025r1v, zhang2025r1vllearningreasonmultimodal, huang2025visionr1incentivizingreasoningcapability, liu2025segzeroreasoningchainguidedsegmentation, deng2025openvlthinker, wang2025vlrethinkerincentivizingselfreflectionvisionlanguage, wang2025sotalessmctsguidedsample} has focused primarily on verifiable rewards applicable to strictly verifiable settings (e.g., geometry mathematics \citep{chen-etal-2021-geoqa}). Incorporating a computationally intensive judge into RL training without incurring substantial cost is non-trivial, and we hope our method and open-source framework will encourage broader consideration of semi-verifiable tasks in vision–language domains.
\newpage

\section{ThinkLite-VL}
As we mention in the Sec. 4.1,  we experiment beyond specific table and chart tasks to more general reasoning, utilizing \textbf{ThinkLite-VL-70K} dataset from recent multimodal reasoning work~\citep{wang2025sotalessmctsguidedsample}, which includes multimodal reasoning tasks from multimodal reasoning \citep{lu2021intergpsinterpretablegeometryproblem, chen2022geoqageometricquestionanswering,seo-etal-2015-solving}, natural image understanding \citep{kahou2018figureqaannotatedfiguredataset,lu2022learn,marino2019okvqavisualquestionanswering}, and chart interpretation \citep{lu2022iconqanewbenchmarkabstract,lu2023dynamicpromptlearningpolicy}. Although \textbf{SVR-R1} demonstrates improvements over standard GRPO training, we were unable to achieve the absolute high accuracy reported in~\citep{wang2025sotalessmctsguidedsample} in standard GRPO training, because reproducible training and evaluation have not yet been fully released. 

\subsection{Full Prompt for VLM Rollout and Inference}
For a fair comparison with the original paper \citep{wang2025sotalessmctsguidedsample}, we use the same prompt for VLM inference and rollout during RL training, following the standard RL-VLM setup~(\cref{fig:mcts}). The prompt format clearly separates the reasoning steps from the final answer, with the answer presented in a boxed layout.
\begin{figure}[htbp]
    \centering
    \begin{tcolorbox}[
        halign=flush left,
        breakable,
        colback=teal!5!white,
        colframe=teal!75!black,
        title={\footnotesize\textbf{Initial Prompt Head, for VLM Inference or Rollout}},
        fonttitle=\bfseries,
        fontupper=\ttfamily,
        width=\columnwidth, 
        enhanced jigsaw,
        boxrule=0.5pt,
        left=1pt, right=1pt, top=1pt, bottom=1pt
    ]
    \scriptsize
    {\color{red}<Image>} {\color{red}<Question>}

    {\color{red}User:}  

    {\color{blue}You FIRST think about the reasoning process as an internal monologue and then provide the final
answer. The reasoning process MUST BE enclosed within <think> </think> tags. The final
answer MUST BE put in /box.}  

    {\color{red}Assistant:}  
    \end{tcolorbox}
    \vspace{-10pt}
    \caption{Full Prompt for VLM Inference or Rollout for ThinkLite-VL Dataset.}
    \vspace{-10pt}
    \label{fig:mcts}
\end{figure}
\subsection{Experiment}
\noindent \textbf{Dataset Split: 70K Split.} The training set comprises 70k examples spanning Geometry3K, GeoQA, and GEOS (math questions with image input) \citep{lu2021intergpsinterpretablegeometryproblem, chen2022geoqageometricquestionanswering, seo-etal-2015-solving}; FigureQA, ScienceQA, and OK-VQA for image understanding \citep{kahou2018figureqaannotatedfiguredataset, lu2022learn, marino2019okvqavisualquestionanswering}; and IconQA and TabMWP for chart understanding \citep{lu2022iconqanewbenchmarkabstract, lu2023dynamicpromptlearningpolicy}.

\noindent \textbf{Difficult Data Selection: 11K Split.} A key contribution of \citep{wang2025sotalessmctsguidedsample} is that sample difficulty critically influences RFT effectiveness. They employ Monte Carlo Tree Search (MCTS) to select hard examples for VLM RL training, constructing an 11k high-difficulty subset from the full 70k dataset and reporting improved sample efficiency and higher test accuracy, compared to training in 70K split.
\begin{figure}[t]
  \centering
  \includegraphics[width=0.9\linewidth]{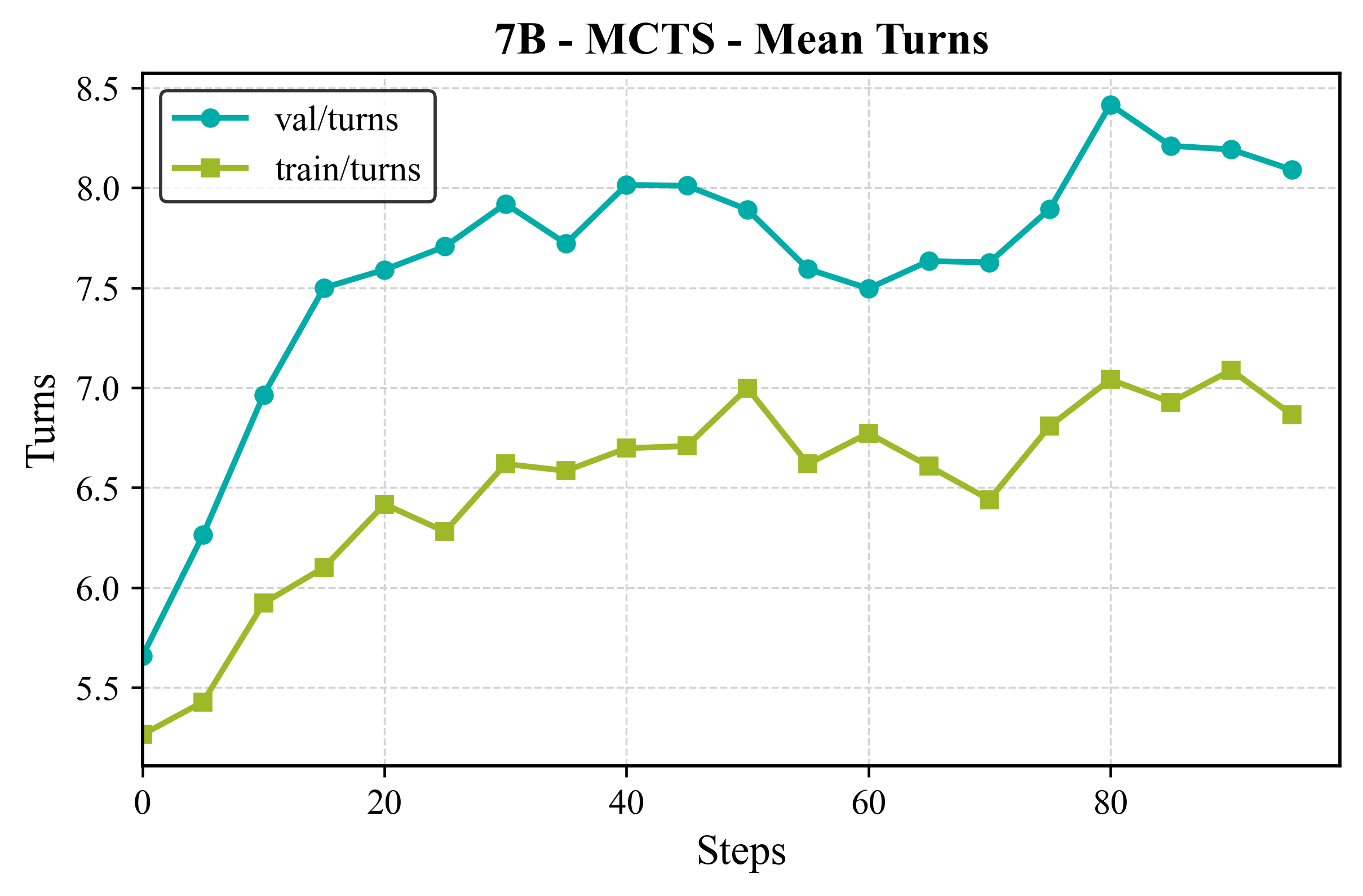}
  \caption{\textbf{Mean number of Turns} vs. \textbf{Training Steps} on train/val splits. We allow a maximum of five verification turns, counting each generation and verification round.}
  \label{fig:thumb-diff}
  \vspace{-1em}
\end{figure}

\noindent \textbf{Experiment Setup.} For the Thinking-VL dataset, we strictly follow the hyperparameter configuration in \citep{wang2025sotalessmctsguidedsample}, which adheres to the EasyR1 default RL-VLM setup. We use the AdamW optimizer \citep{adamw} with an initial learning rate of \(1\times 10^{-6}\). We employ a micro-batch size of 4 per GPU and a mini-batch size of 128 per update, with an overall training batch size of 512. We train Qwen-2.5-VL 7B model on 8×8 A100 GPUs (80GB). We use bf16 precision and set the decoding temperature to 1.0 during rollouts to encourage exploration. The rollout group size is 32. We include a KL-divergence term in the loss with coefficient \(\beta = 1\times 10^{-2}\), larger than that used for table and chart training. We set the maximum number of self-verification rounds to 3 for the 70k split and 5 for the difficult 11k split. We enable asynchronous rollouts for \textbf{SVR-R1}’s multi-turn training and use a dynamic batch size for higher efficiency. For the reward verifier, we follow the authors’ implementation and use mathruler , assigning 0.1 to reward for format (including thinking steps) and 0.9 for the outcome.
\subsection{Supplementary Findings}

\noindent \textbf{SVR-R1 fails to outperform when trained on the 11K difficult split}. As mentioned in the paper, repeatedly rethinking questions that are \textit{too difficult} contributes little to training, as the correct answer may remain unreachable even with unlimited self-verification turns. When training on the 11K difficult split selected by MCTS, we observe no improvement in reasoning accuracy, while the number of verification turns grows dramatically during RL training (\cref{fig:thumb-diff}). This contrasts with the 70K dataset, where we observe decreasing turns and better reasoning performance during training, consistent with the results and findings shown in the table and chart experiments.

\subsection{Implementation Details}
\label{sec:details}
We train each model using the AdamW optimizer \citep{adamw} with an initial learning rate of $
1\times 10^{-6}$. Given the large image input size and long text prompts (up to 16k tokens), we employ a micro-batch size of 2 per GPU and mini-batch size of 256 (for chart) or 128 (for table) for one update. The 3B and 7B model are trained on 8×8 A100 GPUs with 80GB memory, We use bf16 precision and set the decoding temperature to 1.0 during rollouts for exploration. We set the rollout group size to 16. We enable KL-divergence term in loss and set the co-efficient $\beta$ to be $1\times 10^{-3}$. We manually set the maximum self-verification round to 3. 
We enable asynchronous rollouts for \textbf{SVR-R1}'s multi-turn training as \textbf{SVR-R1}'s multiple self-verification loops within each rollout may lead to inefficient GPU utilization if processed strictly synchronously. For additional experiments on \textbf{ThinkLite}, we strictly follow the setup from \citep{wang2025sotalessmctsguidedsample}, with full details provided in the supplementary materials.

\subsection{Computation}
\noindent \textbf{Limited Computational Overhead}. Generation and verification reuse the same model parameters, so modern frameworks do not need to move weights to or from the GPU, resulting in minimal additional computational overhead, only 10\% wall clock time if  properly set up.

\begin{figure}[h]
  \centering
  \includegraphics[width=0.48\linewidth]{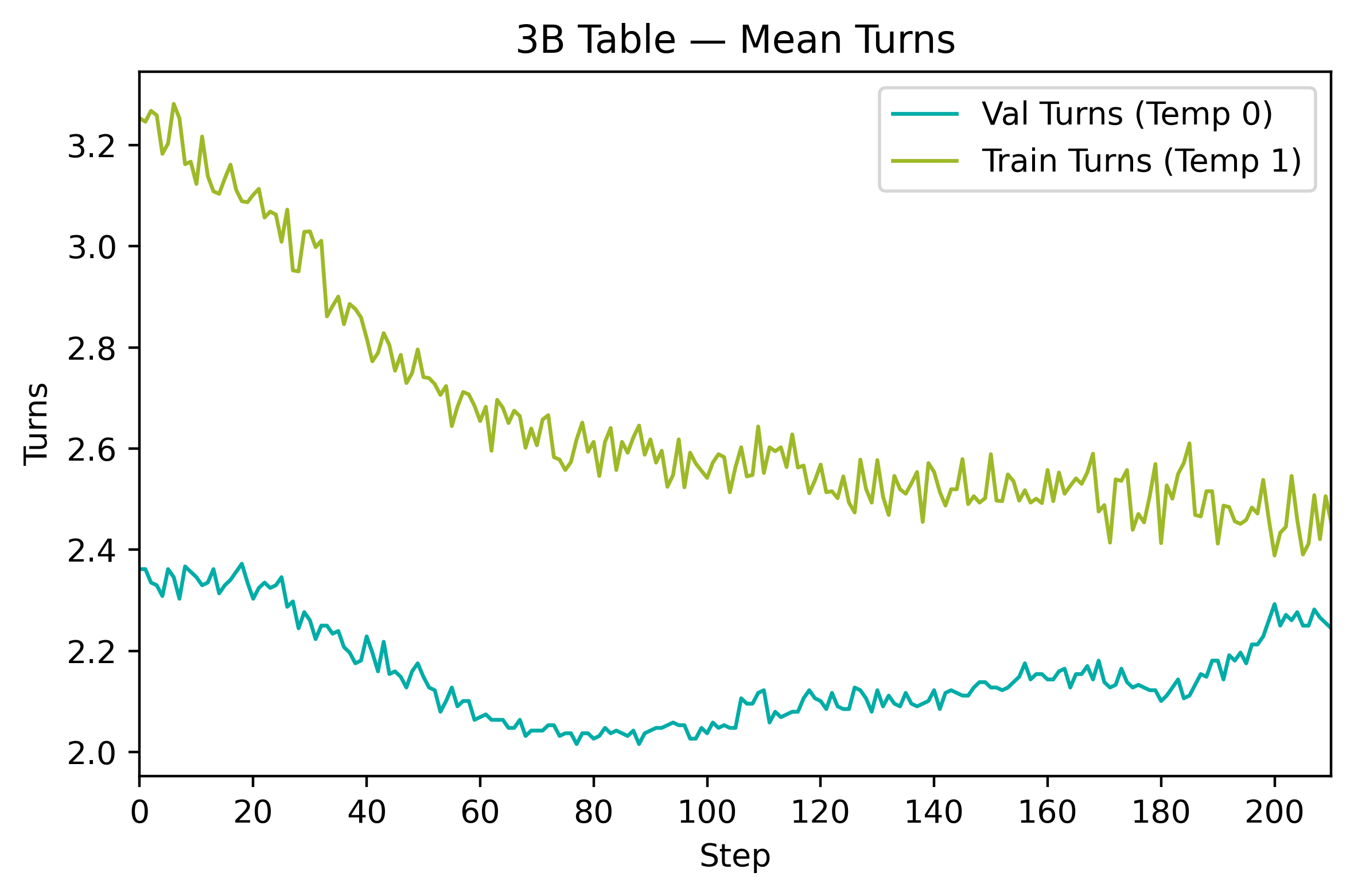}
  \vspace{-0.5em}
  \caption{SVR-R1 surpasses standard GRPO on table tasks. \textbf{Average Turns} vs. \textbf{Training Steps}.}
  \vspace{-1.0em}
  \label{fig:table-turns}
\end{figure}

\end{document}